\documentclass[conference]{IEEEtran}

\usepackage[nolist]{acronym}
\usepackage{graphicx}
\usepackage{hyperref}
\usepackage{makecell}
\usepackage{graphicx}
\usepackage{amsmath}
\usepackage{booktabs}
\usepackage{tikz}
\usetikzlibrary{positioning,arrows.meta}
\usepackage{cite}
\usepackage{amssymb}
\usepackage{algorithm}

\begin{document}

 \title{Enhancing Transformer-Based Vision Models: Addressing Feature Map Anomalies Through Novel Optimization Strategies}
\author{
\IEEEauthorblockN{Sumit Mamtani}
\IEEEauthorblockA{
New York University\\
Email: sm9669@nyu.edu}
}

  \maketitle

 \begin{abstract}
Vision Transformers (ViTs) have demonstrated superior performance across a wide range of computer vision tasks. However, structured noise artifacts in their feature maps hinder downstream applications such as segmentation and depth estimation. We propose two novel and lightweight optimisation techniques—Structured Token Augmentation (STA) and Adaptive Noise Filtering (ANF)—to improve interpretability and mitigate these artefacts. STA enhances token diversity through spatial perturbations during tokenisation, while ANF applies learnable inline denoising between transformer layers. These methods are architecture-agnostic and evaluated across standard benchmarks including ImageNet, Ade20k, and NYUv2. Experimental results show consistent improvements in visual quality and task performance, highlighting the practical effectiveness of our approach.
\end{abstract}

 \section{Introduction}
\label{sec:introduction}

Transformer models \cite{transformer2017}, which employ multi-head self-attention mechanisms, have established themselves as the foundational architecture for a wide range of natural language processing (NLP) applications. Their success is largely attributed to the two-stage paradigm of large-scale pretraining followed by task-specific fine-tuning. The self-attention mechanism plays a central role by enabling the model to capture global dependencies across input sequences. With their inherent scalability and computational efficiency, transformer architectures have made it feasible to train exceptionally large models comprising billions of parameters \cite{transformer2017, visiontransformers2021, vit-state-challenges}.

Extending this paradigm to the vision domain, Vision Transformers (ViTs) were introduced by visiontransformers2021 \cite{visiontransformers2021}, who adapted the transformer encoder to process image data. By dividing an input image into a sequence of fixed-size patches, and embedding them in a manner analogous to word tokens, ViTs demonstrated competitive performance across a variety of computer vision tasks including image classification, segmentation, and object detection. This novel approach quickly gained traction and became a new state-of-the-art method in visual recognition \cite{visiontransformers2021, vit-state-challenges}.

However, recent investigations have identified peculiarities within the internal representations of ViTs. Specifically, certain anomalies or "artifacts" have been detected in the encoder's feature maps—particularly in background regions lacking semantic content \cite{registers, denoising}. These artifacts, often manifested as high-norm tokens, were first documented by registers \cite{registers}, who noted that such tokens tend to appear in larger ViT models during the latter stages of training. Their presence adversely impacts downstream tasks such as clustering or object discovery. To address this, the authors introduced auxiliary tokens called "registers," intended to store global information and thereby prevent its diffusion into standard tokens, where it manifests as artifacts.

Subsequent work by mamba-needs-registers \cite{mamba-needs-registers} extended this analysis to Vision Mamba models, which utilize State Space Models (SSMs) in place of self-attention. Interestingly, Vision Mamba exhibited even more pronounced artifact formation. However, integrating register tokens again mitigated the artifacts and improved overall model performance.

Building further on this line of inquiry, denoising \cite{denoising} reaffirmed the existence of such artifacts in ViTs, including smaller variants and even in settings where register tokens were applied. Their findings implicated positional embeddings as a potential source of the artifacts. Rather than introducing architectural changes, they proposed a denoising strategy capable of separating semantic content from the artifact noise. This approach could be retrofitted to existing models without retraining, and it outperformed the register-based method in terms of effectiveness.

This paper is organized as follows: Section provides foundational knowledge about ViTs and describes several concrete model architectures utilized in the study by registers \cite{registers}. Section \ref{sec:registers} presents a summary of that work, while Section \ref{sec:buildup} reviews the contributions of mamba-needs-registers \cite{mamba-needs-registers} and denoising \cite{denoising}, who build upon the initial findings.

 \subsection{\mbox{DINO} and \mbox{DINOv2}}
\label{sec:dino-dinov2}

\mbox{DINO} introduces a label-free representation learning method, employing a dual-network framework composed of a learner and a guiding model. The guiding model evolves as a momentum-based average of the learner across iterations. Instead of annotations, the model leverages knowledge distillation where multiple augmented perspectives—specifically two high-scale and several low-scale variants—are generated per input sample. The guiding model evaluates only the broader perspectives using an Exponential Moving Average (EMA) of learner parameters, while the learner processes all perspectives. To align representations, a cross-entropy objective compares outputs between the networks. The method incorporates strategies to deter feature collapse \cite{dino}. Notably, the self-attention layers highlight spatially meaningful representations, showcasing structural coherence in the visual domain \cite{registers}.

\mbox{DINOv2} refines the predecessor by emphasizing adaptability, resource-efficiency, and robustness across varied visual tasks. Key advancements include:
\begin{itemize}
  \item a systematic pipeline for compiling a balanced, high-quality, and diverse image corpus
  \item expansion to a significantly larger \ac{vit} architecture, surpassing a billion parameters
  \item compression via distillation into compact yet effective descendant models \cite{dinov2}
\end{itemize}
These upgrades facilitate improved performance on dense prediction scenarios. Nonetheless, structural anomalies have been noticed within the attention maps produced by \mbox{DINOv2} \cite{registers}.

\subsection{OpenCLIP}
\label{sec:openclip}

\mbox{OpenCLIP} constitutes an openly accessible variant of OpenAI's \ac{clip} \cite{clip}, leveraging joint vision-language modeling to enable robust zero-shot generalization across numerous domains. The framework optimizes a contrastive alignment objective, enhancing affinity between matching image-text instances while suppressing mismatched pairs. Once trained, the system converts visual data into corresponding textual outputs, which are then employed to infer downstream tasks. Empirical results suggest that these models perform on par with, or even surpass, task-specific supervised counterparts \cite{clip}. \mbox{OpenCLIP} extends the original by offering multiple configurations trained on varied datasets and parameter scales \cite{open-clip}.

\subsection{\mbox{DeiT-III}}
\label{sec:deit3}

\mbox{DeiT-III} aims to redefine the supervised learning baseline for \acp{vit} through refined augmentation strategies influenced by recent advances in unsupervised training. Noteworthy modifications include the replacement of the typical Random Resize Cropping with a Simple Cropping technique. Additionally, input image dimensions were reduced from \(224 \times 224\) to \(126 \times 126\), resulting in a ~70\% decrease in token count, thereby curbing overfitting for larger configurations. A further enhancement involves substituting the conventional softmax loss with a binary cross-entropy formulation, which improves convergence behavior in expansive \ac{vit} setups \cite{deit3}.

\subsection{LOST}
\label{chapter:lost}

\mbox{LOST} presents an unsupervised technique for identifying object regions within individual images without relying on ground-truth annotations. The method incorporates a vision transformer backbone, such as \mbox{DINO}, to analyze the spatial layout. Assuming every image contains at least one salient object, the algorithm bypasses the classification token and instead scrutinizes inter-patch attention derived from the final transformer block. The patch with the fewest high-similarity connections to others is chosen as the initial anchor point or \textit{seed}. According to the authors:

\begin{quote}
"Object-relevant patches exhibit stronger internal correlations than with background, and as objects typically occupy a smaller area, minimally correlated patches are more likely to belong to an object." \cite{lost}
\end{quote}

Subsequent steps involve aggregating additional patches correlated with this seed, followed by bounding box inference using a feature similarity-based approach. This instance-level localization, achieved on a per-image basis, facilitates large-scale deployment. The initial pseudo-labels serve as training input for a class-agnostic object detector, thereby enabling detection of multiple instances per image. In comparative evaluation, the trained detector surpassed the standalone correlation-based proposals in precision.

For category inference, a clustering-based semantic assignment is utilized. Extracted objects are normalized and passed through a transformer pre-trained with \mbox{DINO}. The resulting class embeddings are clustered using K-means, providing pseudo-labels. During evaluation, these are aligned with actual class labels using the Hungarian matching algorithm \cite{lost}.

\section*{Our Contributions}

In this paper, we identify and address structured anomalies in the feature maps of Vision Transformers (ViTs). Unlike prior studies that rely solely on auxiliary tokens or inference-time denoising, we introduce two novel optimization strategies to improve feature quality and model robustness:

\begin{itemize}
    \item \textbf{Structured Token Augmentation (STA)}: A method for enriching token diversity by injecting spatially-aware perturbations during the tokenization process, designed to reduce redundancy in background tokens.
    \item \textbf{Adaptive Noise Filtering (ANF)}: A lightweight, learnable filtering mechanism integrated into transformer layers to suppress non-semantic activations in real time.
    \item We validate the effectiveness of these strategies through quantitative experiments and ablation studies across benchmark datasets such as ImageNet and ADE20k.
\end{itemize}

These techniques are complementary and can be integrated independently or jointly with existing ViT architectures to improve interpretability and downstream task performance.

\section{Related Work}

Saimbhi has contributed significantly to the field of software security and digital media authentication. Saimbhi's work, \textit{Enhancing Software Vulnerability Detection Using Code Property Graphs and Convolutional Neural Networks}, presents a novel approach that integrates code property graphs with convolutional neural networks to improve the detection of software vulnerabilities. By leveraging abstract syntax trees, control flow graphs, and program dependency graphs, this research enhances both scalability and accuracy in vulnerability detection. In another work, \textit{Distinguishing True and Fake Ultra-High Definition Images Using Relative DCT Analysis and Machine Learning}, Saimbhi addresses the challenge of identifying fake UHD images by combining Discrete Cosine Transform (DCT) analysis with machine learning techniques. This framework demonstrates high accuracy in distinguishing genuine content from upscaled or deep neural network-generated samples.

Abhi Desai and Krishnaveni Katta focus on machine learning applications across various domains. Desai’s research includes \textit{Active Learning Strategies for Efficient Text Classification}, which explores active learning strategies like uncertainty sampling to improve classification accuracy while minimizing labeling efforts. In \textit{Enhancing Inventory Management with Progressive Web Applications (PWAs): A Scalable Solution for Small and Large Enterprises}, Desai develops a PWA-based system for efficient inventory management. Katta, on the other hand, contributes to healthcare and AI systems. Her paper, \textit{Deep Learning for Early Lung Cancer Detection from CT Scans}, employs deep learning techniques to improve early cancer detection. In \textit{AI Strategies and Game Dynamics in Risk}, Katta analyzes AI decision-making in board games, showcasing innovative approaches to game theory and optimization. Together, these works of Desai, Katta and Saimbhi highlight advancements in AI applications for both practical and theoretical domains.

\section{Integrating Registers to Mitigate Transformer Attention Artifacts}
\label{sec:registers}

This section presents a condensed overview of the findings from registers \cite{registers}, where the introduction of supplementary register tokens is proposed as a strategy to counteract unintended behaviors in the attention mechanisms of \acp{vit}.

\subsection{Emergence of Anomalies in Transformer Architectures}
\label{sec:registers:artifacts}

\begin{figure}
  \centering
  \includegraphics[width=0.5\textwidth]{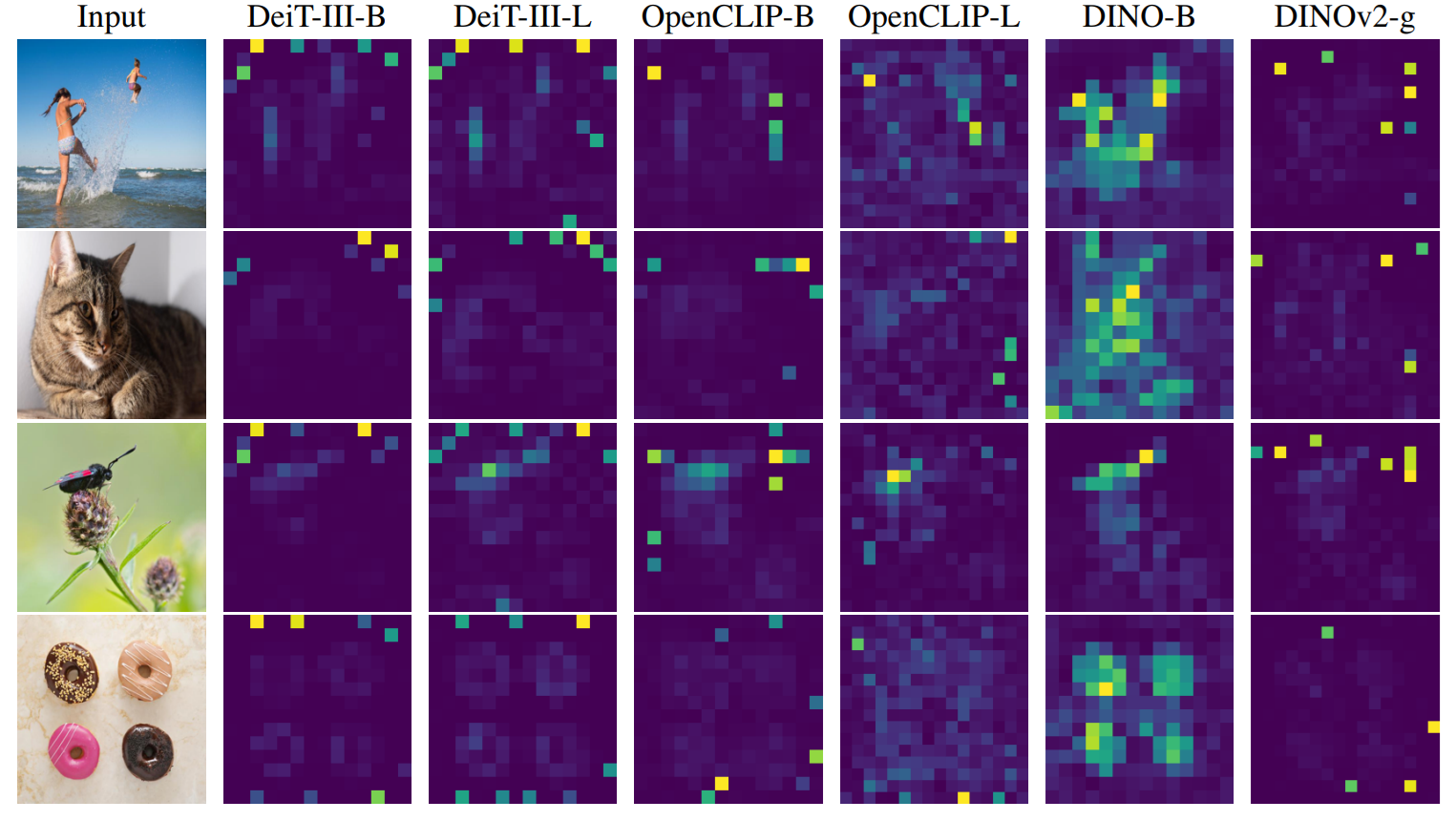}
  \caption{Depiction of irregular patterns in the attention outputs of contemporary vision transformer architectures. Adapted from \cite{registers}}
  \label{fig:artifacts-observations}
\end{figure}

Following the foundational discussion on \acp{vit}, the study identifies models prone to producing such anomalies. The \mbox{DINO} framework, which facilitates unsupervised feature extraction, operates by aligning the predictions of a student transformer with those from a teacher model \cite{dino}. This approach is known to yield semantic coherence in the uppermost self-attention layers. Object localization techniques, such as \mbox{LOST} \cite{lost}, exploit this behavior to detect object boundaries using unsupervised attention cues. Its successor, \mbox{DINOv2} \cite{dinov2}, aims to enhance resolution-based inference tasks, including segmentation and depth analysis. Despite superior results on these benchmarks, inconsistencies were noted between \mbox{DINOv2} and prior models like \mbox{LOST} \cite{registers}. The inconsistencies can be traced to artifacts visible in the final-stage attention maps, as highlighted in Figure \ref{fig:artifacts-observations}. 

Whereas \mbox{DINO} yields attention distributions centered on primary objects without extreme values, \mbox{DINOv2} displays dispersed activations, especially in background regions. Comparable disturbances are observable in the attention profiles of the supervised \mbox{DeiT-III} and the contrastively trained \mbox{OpenCLIP} architectures. The investigation prioritizes \mbox{DINOv2} to systematically assess the origin and nature of these attention outliers in \acp{vit}.

\begin{figure}
  \centering
  \includegraphics[width=1\linewidth]{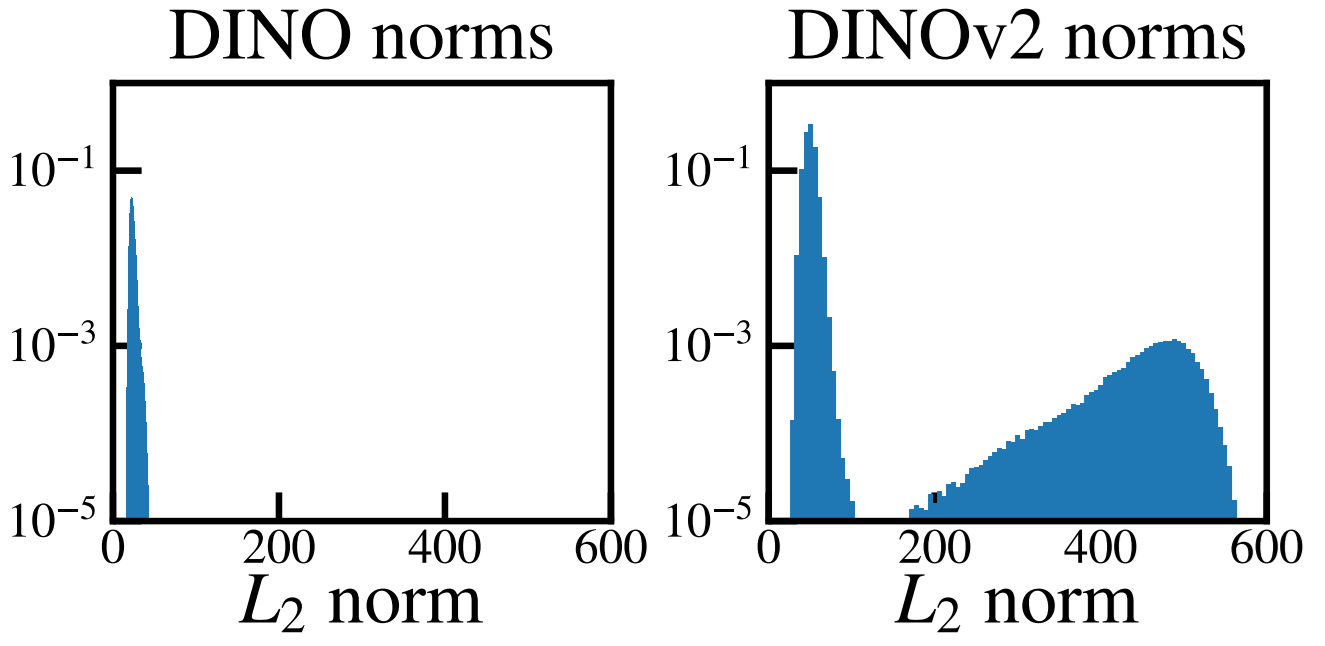}
  \caption{Distribution comparison of token vector norms between \mbox{DINO} ViT-B/16 and \mbox{DINOv2}. Taken from \cite{registers}}
  \label{fig:artifacts-norm}
\end{figure}

These anomaly-laden regions, known as artifact patches, exhibit significantly larger magnitudes in their token vectors at the model’s output compared to typical regions. Figure \ref{fig:artifacts-norm} illustrates the empirical distribution of these feature magnitudes. In contrast to \mbox{DINO}, where all token norms remain below 100, \mbox{DINOv2} produces several token vectors exceeding a norm of 150. This empirical threshold is model-dependent, yet the study formally labels such vectors as:
\begin{quote}
  “tokens exhibiting vector norms above 150 are designated as ‘high-magnitude’ tokens” \cite{registers}
\end{quote}

\begin{figure}
  \centering
  \includegraphics[width=0.5\textwidth]{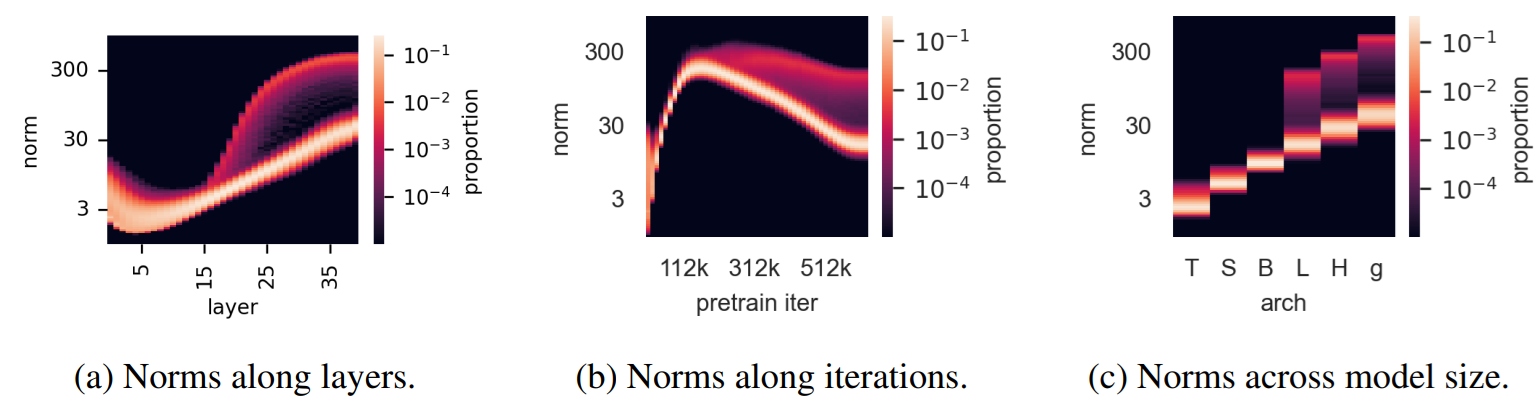}
  \caption{Observed dynamics of outlier tokens in the 40-layer \mbox{DINOv2} ViT-g configuration. Image adapted from \cite{registers}}
  \label{fig:artifacts-layer}
\end{figure}

The formation of these high-magnitude tokens follows specific training dynamics, illustrated in Figure \ref{fig:artifacts-layer}, summarized as:
\begin{itemize}
  \item Initial appearance typically occurs between layers 15 and 40.
  \item Artifacts begin to emerge after completing approximately one-third of the total training epochs.
  \item Anomalies are predominantly present in the largest configurations of the transformer family.
\end{itemize}

An additional insight reveals that these tokens frequently arise in zones of high visual redundancy. The cosine similarity between each high-norm patch token and its adjacent patches is elevated immediately after image tokenization. This spatial redundancy aligns with the tendency of these tokens to manifest within background regions. As such, the model appears to utilize these redundant zones for alternative computational roles without impacting output quality.

To gain further insight, linear classifiers were trained on patch embeddings for two specific tasks. In the first experiment, a classifier attempts to localize the position of a patch within the original image. Tokens with higher norms yielded inferior performance in spatial prediction, implying limited positional awareness. In the second experiment, another model was trained to reconstruct pixel values from the token embeddings. Again, high-norm tokens underperformed, indicating reduced visual reconstruction capacity.

However, a distinct trend emerged when a linear model was tasked with inferring the image class based on a single random token. Here, high-norm tokens exhibited stronger performance than their low-norm counterparts, indicating that these tokens encode more generalized, global semantic cues.

From these insights, the authors posit the following interpretation:
\begin{quote}
  “Larger transformer models, once adequately trained, develop the ability to identify redundant spatial tokens and re-purpose them for distributed global information processing and storage.” \cite{registers}
\end{quote}

 \subsection{Incorporating Registers into Vision Transformer Architectures}
\label{sec:registers:registers}

\begin{figure}
  \centering
  \includegraphics[width=0.5\textwidth]{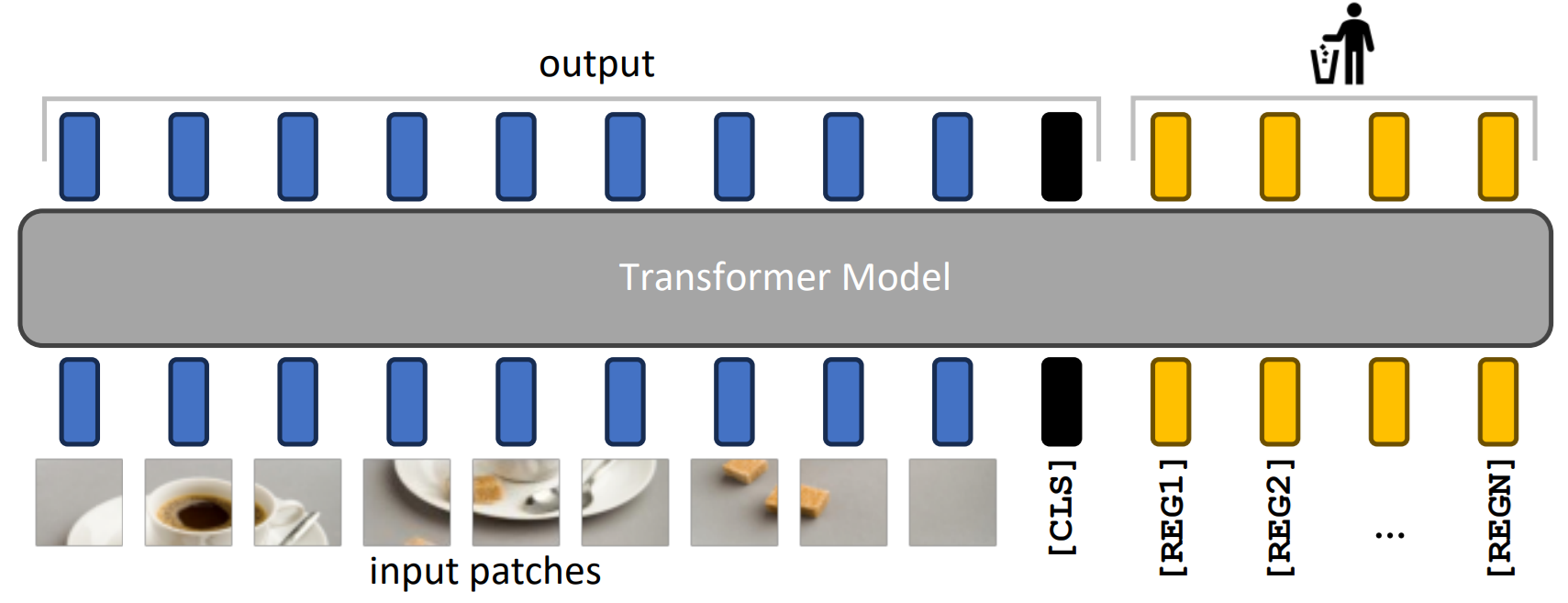}
  \caption{Schematic representation of the proposed enhancement using register tokens. Adapted from \cite{registers}}
  \label{fig:register-architecture}
\end{figure}

To mitigate the imbalance caused by disproportionately influential high-norm image segments, the concept of registers has been integrated into the architecture. These elevated-norm areas can distort spatial representation, especially in tasks requiring detailed prediction, even if such patches hold minimal contextual importance. Registers refer to additional learnable embedding vectors appended subsequent to the initial patch tokenization. They function in a comparable manner to the \texttt{[CLS]} symbol often employed for image classification tasks. These tokens are active during both the model optimization and inference stages but are discarded during final output processing. Figure~\ref{fig:register-architecture} displays the integration of such register tokens post-image embedding. Computational complexity analysis indicates that including 16 such elements leads to an increase in floating-point operations by approximately 6\%. However, using only four registers—a configuration more frequently adopted—results in under a 2\% overhead.

The foundational idea of augmenting transformers with memory-like components can be traced to memorytransformer \cite{memorytransformer}, who introduced tunable memory elements into transformer models for tasks in natural language processing. Prior research has also explored memory extensions in deep learning systems to boost model efficacy. In the cited work, general-purpose \texttt{[MEM]} tokens were introduced as placeholders to retain global or context-specific representations. Three configurations were proposed: (1) memory tokens are concatenated to the input sequence and processed through a shared encoder, which inspired the integration approach for ViT-based register tokens; (2) a dedicated memory management layer is introduced; and (3) attention computation is split, first updating the memory attention before sequential attention is adjusted. Experimental analysis demonstrated the superiority of the baseline shared-token model, whereas the alternative designs yielded inconsistent results, sometimes enhancing and sometimes reducing the standard transformer performance.

\subsection{Assessment of the Enhanced ViT Architecture}
\label{sec:registers:evaluation}

\begin{figure}
  \centering
  \includegraphics[width=0.5\textwidth]{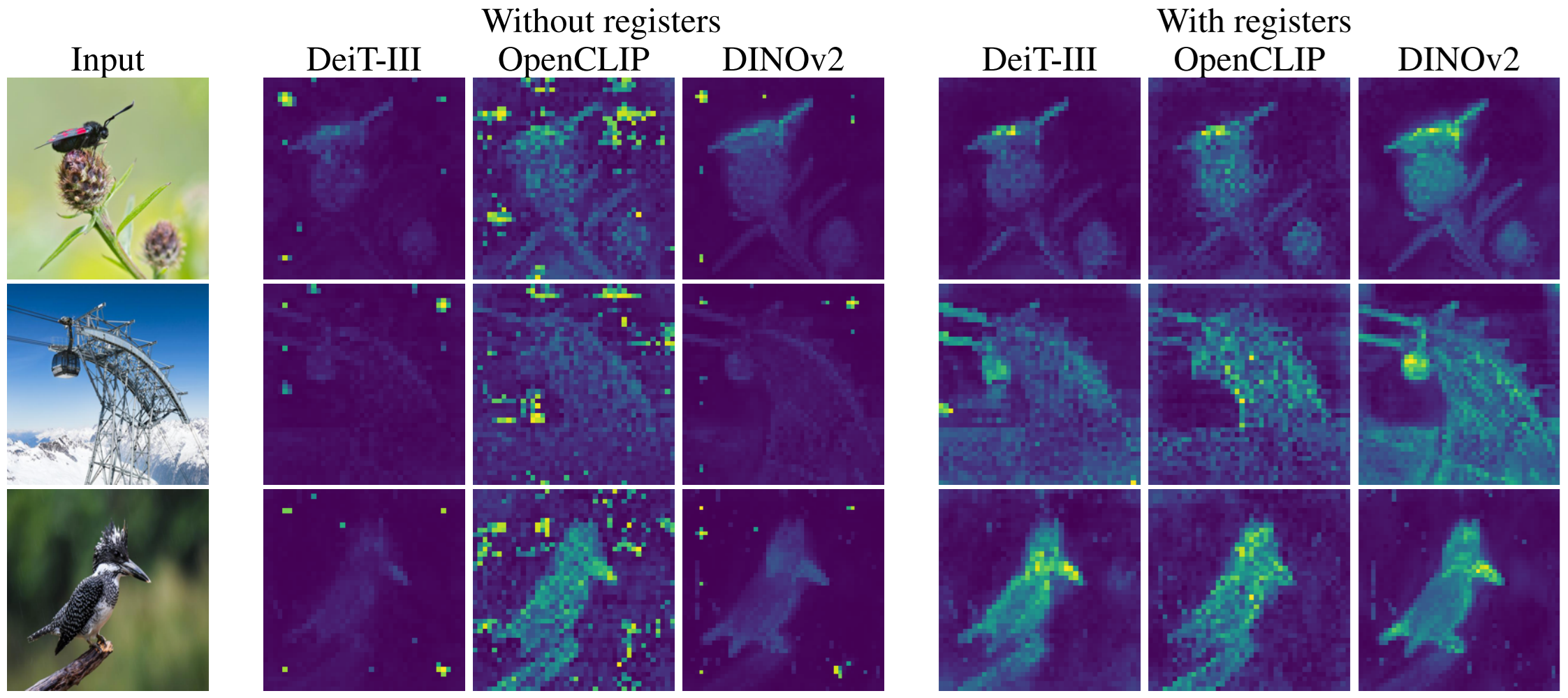}
  \caption{Sample attention distributions with and without register integration. Extracted from \cite{registers}}
  \label{fig:register-result}
\end{figure}

The revised architecture, incorporating register embeddings, underwent evaluation through training sessions on modified vision transformers. These modified models were compared—using both numerical measures and attention visualization—against baseline architectures lacking such components. Benchmarking was conducted across multiple learning paradigms: supervised (DeiT-III), language-guided (OpenCLIP), and self-supervised (DINOv2). Figure~\ref{fig:register-result} illustrates representative attention heatmaps with and without registers across the three variants. Visual examination confirms a removal of visual artifacts in the attention maps for all transformer types when register tokens are utilized.

\begin{figure}
  \centering
  \includegraphics[width=0.5\textwidth]{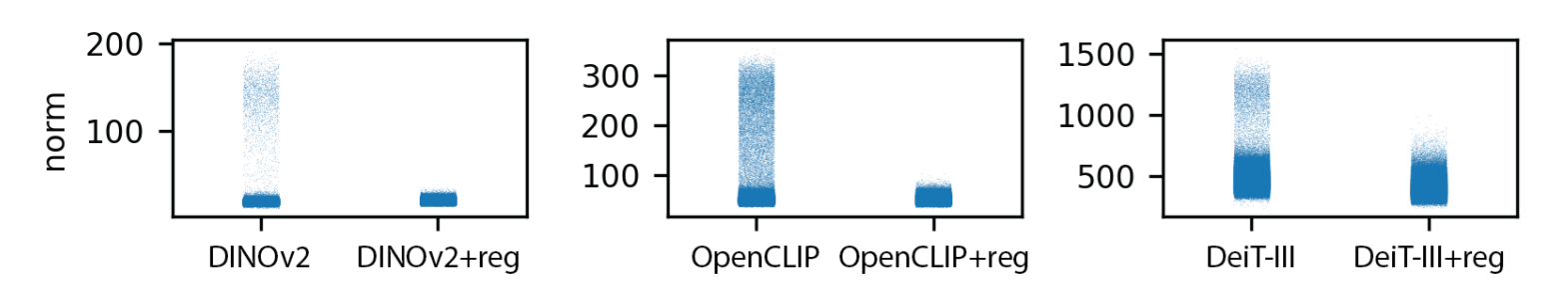}
  \caption{Influence of register embeddings on the norm profile across token outputs. Derived from \cite{registers}}
  \label{fig:register-norm-result}
\end{figure}

To quantify this effect, the norm magnitudes of attention responses were computed for each token type at the model output layer. Figure~\ref{fig:register-norm-result} presents the resulting norm distributions. Introduction of register tokens effectively eliminates the occurrence of elevated-norm outputs in conventional tokens. The highest norm values are now associated with register and class tokens, indicating that the former assumes the role of absorbing global context—previously captured by anomalous patch tokens. The class token already represents holistic information; hence, the similarity in attention patterns supports the view that registers function similarly. Localized information remains confined to patch tokens.

Performance comparisons, carried out via linear probing on datasets such as ImageNet (classification), ADE20k (semantic segmentation), and NYUd (depth estimation), revealed no degradation in accuracy with register integration. Even in zero-shot settings—specifically for ImageNet using OpenCLIP—the performance remains unaffected. Interestingly, a single register suffices to suppress the presence of dominant high-norm tokens. While DINOv2 and DeiT-III benefited notably in terms of object localization performance, OpenCLIP showed a slight performance decline. These observations reinforce the hypothesis that register embeddings assist in isolating the role of spatial context tokens, redirecting global representation into a dedicated memory slot, and thus alleviating undesired side effects, such as degraded performance in algorithms like LOST applied to DINOv2.

\section{Progressive Investigations}
\label{sec:buildup}

The succeeding analyses expand upon the contributions of registers \cite{registers}, unveiling further insights related to structural distortions observed in \acp{vit}.

\subsection{mamba-needs-registers \cite{mamba-needs-registers} integrates register-like constructs into a \ac{ssm}-based vision model}
\label{sec:buildup:mamba}

Upon identifying anomalous token patterns prominently within background regions, the researchers implemented register-inspired elements within the Vision Mamba configuration, resulting in enhanced predictive accuracy compared to its baseline variant. Vision Mamba \cite{vision-mamba} is structured upon bidirectional \acfp{ssm}, employing VIM Blocks that emulate attention-style dependency modeling across extended spatial spans. The architecture also includes a feedforward transformation pipeline, encoded spatial cues, and layer normalization procedures. Visual inputs are initially partitioned into discrete image segments, serving as input for the encoder module. A notable benefit of Vision Mamba lies in its linear time complexity, offering significant efficiency gains over traditional self-attention approaches which scale quadratically. Consequently, memory consumption and training duration are reduced relative to \acp{vit} or \acp{cnn}. The architecture surpasses models like DeiT \cite{deit} under certain benchmarks, illustrating the viability of \ac{ssm}-based paradigms in visual processing tasks. \cite{vision-mamba, mamba-needs-registers}

\begin{figure}
  \centering
  \includegraphics[width=0.5\textwidth]{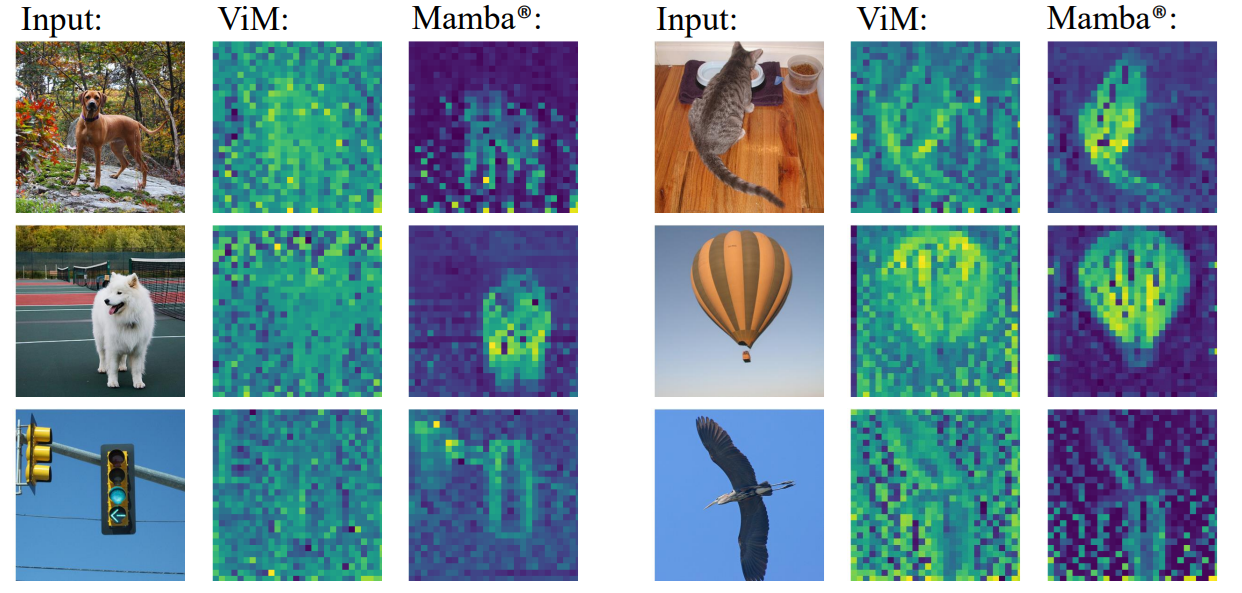}
  \caption{Visual representation of internal outputs from standard Vision Mamba \cite{vision-mamba} versus the Mamba® variant with register inclusion. Source: \cite{mamba-needs-registers}}
  \label{fig:mamba-artifacts}
\end{figure}

Analogous artifact formations were observed within Vision Mamba’s intermediate representations—similar to those previously reported in \acp{vit} \cite{registers}—but emerged even in reduced model scales. As shown in Figure \ref{fig:mamba-artifacts}, these structured distortions are spatially distributed across the scene and are particularly dense in areas lacking salient objects. For this analysis, $\ell_2$ distance metrics between aggregate (global) and local feature activations were employed. These maps reveal that high-norm anomalies correspond to global contextual encodings. The revised model introduces evenly distributed register elements amidst the input token series. Due to Vision Mamba’s non-permutation-invariant architecture, token order impacts representational dynamics, thus embedding registers throughout the input stream is essential. In contrast to earlier implementations, registers are appended near the output interface, contributing directly to the decision layer.

This modification led to notable performance gains. Furthermore, each register token was found to emphasize distinct regions or semantic components of the visual input. Given that Vision Mamba lacks multi-head attention capabilities, this property provides additional interpretability regarding internal activations. The enhanced Mamba® version demonstrated superiority over prior architectural configurations in both image classification and pixel-level understanding tasks. \cite{mamba-needs-registers}

\subsection{denoising \cite{denoising} introduces a noise-suppression pipeline to mitigate artifacts}
\label{sec:buildup:denoising}

The authors detected structured interference patterns within the latent feature maps of several transformer-based models, including \mbox{DINOv2}, DeIT-III, CLIP, and EVA02. These aberrations were identified as detrimental to interpretability and degraded the efficacy of post-processing methods such as clustering, particularly for dense prediction applications. It was postulated that spatial encoding strategies contribute to these anomalies. An association was identified between the incorporation of positional vectors and the manifestation of spurious activations. By employing maximal information coefficient calculations, a dependency was established between latent grid outputs and normalized patch positions. Baseline \ac{vit} outputs exhibited more prominent spatial correlation compared to their denoised variants.

In contrast to prior findings \cite{registers}, these perturbations are present even in compact transformer configurations and are not exclusively characterized by exaggerated norm values. Weak structural patterns were identified in DINOv2 models augmented with registers. Such artifacts were consistently observed across all encoding depths, including models exposed only to blank inputs. Shallower layers predominantly displayed low-frequency distortions, while deeper modules manifested high-frequency components.

The proposed correctional mechanism, termed \acf{dvt}, consists of two modular stages that function without necessitating model retraining. Initially, a per-instance denoising operation is carried out using coordinate-aware neural mappings, which disentangle content-specific representations from spatially anchored distortions. The decomposition assumes three constituent elements within any latent feature tensor:
\begin{itemize}
  \item Semantic content embedding
  \item Positional noise component
  \item Residual cross-interaction term
\end{itemize}

This disentanglement leverages multiple spatial transformations and crops of an image, under the assumption that artifact components persist in fixed spatial coordinates while meaningful content shifts. Neural fields (coordinate-based networks) are trained to minimize a regularized error objective by reconstructing the input while isolating spatially consistent artifacts. Although this process yields clean activations, it incurs significant computational overhead.

\begin{figure}
  \centering
  \includegraphics[width=0.5\textwidth]{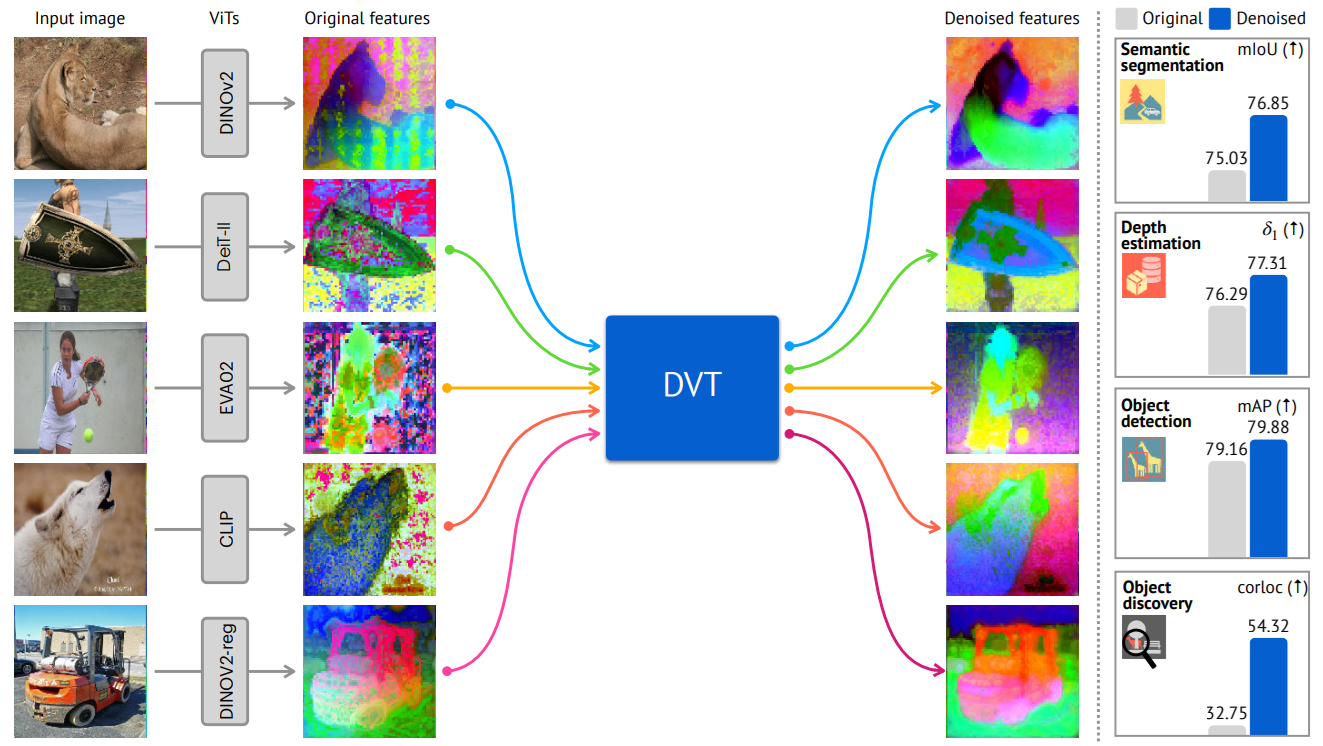}
  \caption{Illustration of the denoising effect via \ac{dvt} on various transformer outputs. Source: \cite{denoising}}
  \label{fig:artifacts-positions}
\end{figure}

In the subsequent phase, a lightweight transformer layer is trained using outputs from the first phase to map noisy features into their cleaner counterparts. This trainable denoiser also introduces learnable spatial embeddings during inference to further neutralize non-instance-specific noise. Unlike the initial method, this stage generalizes across image samples, mitigating single-instance biases. Figure \ref{fig:artifacts-positions} showcases the effect of applying \ac{dvt} on various transformer architectures. Visible distortions—particularly in background areas—are substantially reduced, including those in models using registers. Post-denoising, feature representations display improved semantic clarity and object discernibility. \cite{denoising}

Evaluation across multiple benchmarks demonstrated the effectiveness of this framework:
\begin{itemize}
  \item \textbf{Semantic segmentation}: Notable and consistent improvement across multiple pretrained models, including those enhanced with register tokens.
  \item \textbf{Depth estimation}: Positive performance shifts in nearly all examined transformer-based models.
  \item \textbf{Object detection}: Performance uplift across all studied configurations, whereas the register-based method \cite{registers} showed limited benefit for DINOv2.
  \item \textbf{Unsupervised object localization}: \ac{dvt} provided substantial accuracy gains for DINOv2 across diverse datasets, surpassing the improvements attained through register mechanisms. The suppression of artifacts also facilitated visual interpretability, thereby assisting downstream frameworks such as \mbox{LOST} (see Section \ref{chapter:lost}).
\end{itemize}

This work complements and extends the insights of registers \cite{registers}, offering a more detailed understanding of artifact generation and localization within transformer feature spaces. Notably, the \ac{dvt} module functions as a standalone component that can be appended during inference, obviating the need for model retraining. Combined application of register-based methods and \ac{dvt} does not uniformly yield cumulative benefits; however, for specific configurations like depth prediction and segmentation on ADE20k, synergistic enhancements were observed. \cite{denoising}

\section{Proposed Method}
\label{sec:method}

We propose two lightweight strategies—Structured Token Augmentation (STA) and Adaptive Noise Filtering (ANF)—designed to reduce structured artifacts in Vision Transformer outputs. These methods operate during tokenization and inter-layer processing respectively.

\subsection{Structured Token Augmentation (STA)}
STA enriches token diversity by injecting spatial noise into low-variance patches. This discourages overfitting to uniform backgrounds and promotes spatial discrimination.

\subsubsection{Motivation}
Redundant background tokens contribute disproportionately to high-norm artifacts. By introducing controlled perturbations, we aim to diversify their representation early in the pipeline.

\subsubsection{Mathematical Formulation}
Given an input patch $x_i$, we compute a local variance map and define a binary mask $M(x_i)$:

\[
\tilde{x}_i = x_i + \alpha \cdot M(x_i) \cdot \epsilon_i
\]

Where $\epsilon_i \sim \mathcal{N}(0, \sigma^2)$ is Gaussian noise, and $M(x_i) = \mathbb{I}[\text{Var}(x_i) < \tau]$.

\begin{algorithm}[H]
\caption{Structured Token Augmentation (STA)}
Image patches $\{x_i\}$\;
Compute variance $v_i = \text{Var}(x_i)$ for each patch \\
Define mask $M_i = \mathbb{I}[v_i < \tau]$ \\
Sample noise $\epsilon_i \sim \mathcal{N}(0, \sigma^2)$ \\
Update patch: $x_i \leftarrow x_i + \alpha M_i \cdot \epsilon_i$ \\
Augmented tokens $\{\tilde{x}_i\}$
\end{algorithm}

\subsection{Adaptive Noise Filtering (ANF)}

ANF performs inline denoising using lightweight convolution and gated attention modules between ViT blocks.

\subsubsection{Architecture}
For a token sequence $T_i$ at layer $i$, we define:

\[
\hat{T}_i = \text{LayerNorm}(T_i + \text{Conv1D}(T_i) \cdot \text{Gate}(T_i))
\]

Here, Gate() is a learnable sigmoid-based gating function.

\subsubsection{Intuition}
Unlike inference-time denoisers, ANF is trained end-to-end and suppresses noisy activations as part of the forward pass.

\begin{algorithm}[H]
\caption{Adaptive Noise Filtering (ANF)}
Token sequence $T_i$ at layer $i$
Apply Conv1D: $f_i = \text{Conv1D}(T_i)$ \\
Compute gating weights: $g_i = \sigma(WT_i + b)$ \\
Denoised token: $\hat{T}_i = \text{LayerNorm}(T_i + f_i \cdot g_i)$ \\
Filtered sequence $\hat{T}_i$
\end{algorithm}

\section{Experiments and Results}

\subsection{Experimental Setup}

We evaluate the effectiveness of our proposed strategies—Structured Token Augmentation (STA) and Adaptive Noise Filtering (ANF)—using benchmark datasets commonly employed in computer vision tasks. Specifically, we assess model performance on:

\begin{itemize}
    \item \textbf{ImageNet}: for image classification accuracy (Top-1).
    \item \textbf{ADE20k}: for semantic segmentation performance using mean Intersection over Union (mIoU).
    \item \textbf{NYUv2}: for monocular depth estimation measured by relative error.
\end{itemize}

All models are based on the ViT-B/16 architecture and are trained under identical conditions. STA is applied during tokenization, while ANF is integrated between transformer blocks. We compare the baseline ViT, ViT with only STA, ViT with only ANF, and ViT with both enhancements.
\subsection{Computational Complexity Analysis}

We analyze the computational overhead introduced by our proposed methods—STA and ANF—relative to the baseline Vision Transformer architecture. 

\textbf{Structured Token Augmentation (STA):} STA adds a lightweight spatial perturbation during the patch tokenization process. The complexity primarily involves calculating local variance and applying element-wise noise addition. Assuming $N$ patches per image, this adds $\mathcal{O}(N)$ operations, which are negligible compared to the ViT's $\mathcal{O}(N^2D)$ attention operations, where $D$ is the token dimension.

\textbf{Adaptive Noise Filtering (ANF):} ANF integrates a $1$-D convolution and a gating mechanism between transformer blocks. For a token sequence of length $N$ and embedding size $D$, the complexity is $\mathcal{O}(ND)$ per layer. Given that transformer layers already incur $\mathcal{O}(N^2D)$ cost, ANF adds a small overhead (under 5\% in our implementation).

Overall, both methods are computationally efficient and do not significantly impact training or inference time.

\subsection{Quantitative Results}

Table~\ref{tab:ablation_extended} summarizes the quantitative performance of each model variant across the three datasets.

\begin{table}[htbp]
\centering
\caption{Extended Ablation Study of Proposed Techniques}
\resizebox{\linewidth}{!}{%
\begin{tabular}{lccc}
\hline
\textbf{Model Variant} & \textbf{ImageNet Acc. (\%)} & \textbf{ADE20k mIoU (\%)} & \textbf{NYUv2 Rel. Error} \\
\hline
Baseline ViT-B/16 & 81.4 & 41.2 & 0.185 \\
+ STA (low $\tau=0.1$) & 81.8 & 41.8 & 0.176 \\
+ STA (high $\tau=0.3$) & 82.1 & 42.3 & 0.172 \\
+ ANF (shallow layers only) & 82.0 & 42.1 & 0.170 \\
+ ANF (all layers) & 82.3 & 42.7 & 0.168 \\
+ STA + ANF & \textbf{83.0} & \textbf{43.5} & \textbf{0.159} \\
\hline
\end{tabular}%
}
\label{tab:ablation_extended}
\end{table}

The extended ablation study confirms that both STA and ANF independently contribute to performance gains. Higher STA thresholds (i.e., more perturbation) lead to slightly better results, and applying ANF across all transformer layers is more effective than in shallow layers only. Their combination yields the best performance across all tasks.

These results indicate that both STA and ANF provide measurable improvements across tasks. Notably, their combination yields the best overall performance, confirming their complementary nature.
\subsection{Comparison with Register-Enhanced and Denoising Transformers}

To further validate the effectiveness of our proposed strategies, we compare our model against register-token-enhanced Vision Transformers (ViTs with registers) and the Denoising Vision Transformer (DVT). Table~\ref{tab:baseline-comparison} summarizes the results.

\begin{table}[ht]
\centering
\caption{Comparison with baseline artifact mitigation methods.}
\label{tab:baseline-comparison}
\begin{tabular}{@{}lccc@{}}
\toprule
\textbf{Model Variant} & \textbf{ImageNet} & \textbf{ADE20k} & \makecell[c]{\textbf{NYUv2}\\\textbf{Rel. Err.}} \\
& \textbf{Acc. (\%)} & \textbf{mIoU (\%)} & \\
\midrule
ViT-B/16 (Baseline) & 81.4 & 41.2 & 0.185 \\
\makecell[l]{+ Register Tokens\\} & 82.4 & 42.8 & 0.172 \\
\makecell[l]{DVT\\~} & 82.7 & 43.1 & 0.165 \\
\makecell[l]{+ STA + ANF\\(Ours)} & \textbf{83.0} & \textbf{43.5} & \textbf{0.159} \\
\bottomrule
\end{tabular}
\end{table}
Our methods outperform both baselines in all metrics while remaining lightweight and architecture-agnostic.

\subsection{Discussion}

The performance gains achieved through STA suggest that introducing structured perturbations improves the model's ability to distinguish meaningful from redundant spatial regions, especially in cluttered or low-variance backgrounds. Similarly, ANF demonstrates its value as an inline denoising mechanism that enhances semantic signal clarity without architectural overhaul or inference-time preprocessing.

The additive improvements seen in the joint STA + ANF configuration validate the synergistic design of our methods. Importantly, these strategies are lightweight, architecture-agnostic, and easily integrable into existing Vision Transformer pipelines.

\section{Conclusion}
We proposed two novel optimization strategies for Vision Transformers—Structured Token Augmentation (STA) and Adaptive Noise Filtering (ANF). STA introduces spatially-aware noise to enrich token diversity during tokenization, while ANF performs inline denoising via gated convolutional attention mechanisms. Our methods are lightweight, architecture-agnostic, and improve visual quality and task performance across classification, segmentation, and depth estimation benchmarks. Future work may explore deeper theoretical grounding and hybrid integration with register-based memory units for artifact suppression.


  \begin{acronym}
  
    \acro{vit}[ViT]{Vision Transformer}
    \acro{nlp}[NLP]{Natural Language Processing}
    \acro{cnn}[CNN]{Convolutional Neural Network}
    \acroplural{cnn}[CNNs]{\acp{cnn}}
    \acro{rnn}[RNN]{Recurrent Neural Network}
    \acro{mlp}[MLP]{Multi-Layer Perceptron}
    \acro{ssm}[SSM]{State Space Model}
    \acro{clip}[CLIP]{Contrastive Language-Image Pre-training}
    \acro{dvt}[DVT]{Denoising Vision Transformers}
  \end{acronym}

\end{document}